\documentclass[runningheads]{llncs}
\usepackage{graphicx}
\usepackage{dirtytalk}
\usepackage{hyperref}
\hypersetup{
    colorlinks=true,
    linkcolor=blue,
    filecolor=magenta,      
    urlcolor=cyan,
}
 
\begin{document}
\title{Stories for Images-in-Sequence by using Visual and Narrative Components
\thanks{This research was partially funded by Pendulibrium and the Faculty of computer science and engineering, Ss. Cyril and Methodius University in Skopje.} 
}
%
%
\author{Marko Smilevski\inst{1,2}\and
Ilija Lalkovski\inst{2}\and
Gjorgji Madjarov\inst{1, 3}}
\authorrunning{M. Smilevski et al.}
%
\institute{Ss. Cyril and Methodius University, Skopje, Macedonia \and
Pendulibrium, Skopje, Macedonia \and Elevate Global, Skopje, Macedonia  \\
\email{\{marko.smilevski,ilija\}@webfactory.mk, gjorgji.madjarov@finki.ukim.mk} }
\maketitle              
\begin{abstract}
Recent research in AI is focusing towards generating narrative stories about visual scenes. It has the potential to achieve more human-like understanding than just basic description generation of images-in-sequence. In this work, we propose a solution for generating stories for images-in-sequence that is based on the Sequence to Sequence model. As a novelty, our encoder model is composed of two separate encoders, one that models the behaviour of the image sequence and other that models the sentence-story generated for the previous image in the sequence of images. By using the image sequence encoder we capture the temporal dependencies between the image sequence and the sentence-story and by using the previous sentence-story encoder we achieve a better story flow. Our solution generates long human-like stories that not only describe the visual context of the image sequence but also contains narrative and evaluative language. The obtained results were confirmed by manual human evaluation.

\end{abstract}
\section{Introduction}
Storytelling is one of the oldest and most important activities known to mankind. It predates writing and for a long time, it was the only way to pass knowledge to the next generations. Storytelling is often used as a technique to unfold the narrative of a story, that describes scenes or activities. Storytelling is closely related to sight because visual context and narrative are the core inspiration for stories. Over the last few years, the improvements in computer vision have enabled the machines to \say{see} and generate labels about given images. This has allowed researchers to gain significant progress in the field of image captioning, whose goal is to generate a description for a given image, and video sequence description, which is generating a description for a sequence of images. The next degree of reasoning, in terms of generating text from a sequence of images, is generating a narrative story about a sequence of images, that depicts events that are happening consequently. This means that we shift our focus from generating a description, to generating a story. This problem in machine learning is also known as visual storytelling. Storytelling is more complex than plain description because it uses more abstract and evaluative language for the activities in the images. This means that our goal is to generate sentences like \say{They enjoyed their dinner} instead of \say{Four people are sitting by the table}.

We introduce a novel solution that is based on the Sequence to Sequence~\cite{DBLP:journals/corr/SutskeverVL14} model. The model generates stories, sentence by sentence with respect to the sequence of images and the previously generated sentence. The architecture of our solution consists of an image sequence encoder that models the sequential behaviour of the images, a previous-sentence encoder and a current-sentence decoder. The previous-sentence encoder encodes the sentence that was associated with the previous image and the current-sentence decoder is responsible for generating a sentence for the current image of the sequence. We also introduce a novel way of grouping the images of the sequence during the training process, in order to encapture the effect of the previous images in the sequence. Our goal with this approach was to create a model that will generate stories that contain more narrative and evaluative language and that every generated sentence in the story will be affected not only by the sequence of images but also by what has been previously generated in the story. 

\section{Related work}
In the last couple of years, research in the domain of vision to language has grown exponentially. In particular, research is divided into three sub-categories: Description of images-in-isolation, Description of images-in-sequence and Story for images-in-sequence.
\subsection{Description of images-in-isolation}
Description of images-in-isolation is the problem of generating a textual description for an image. This category is represented by image captioning. Current caption generation research focuses mainly on concrete conceptual image descriptions of elements directly depicted in a scene~\cite{framing_image_descriptions}. Image captioning is a task whose input is static and non-sequential (an image rather than, say, a video), whereas the output is sequential (a multi-word text), in contrast to non-sequential outputs such as object labels~\cite{DBLP:journals/corr/DonahueHGRVSD14}. An extensive overview of the datasets available for image captioning is provided by~\cite{DBLP:journals/corr/BernardiCEEEIKM16}. The three biggest datasets are MS COCO~\cite{DBLP:journals/corr/LinMBHPRDZ14}, SBU1M Captions~\cite{NIPS2011_4470}, Deja-Image Captions~\cite{N15-1053}. 
Work done by~\cite{DBLP:journals/corr/KarpathyF14} and~\cite{DBLP:journals/corr/VinyalsTBE14} has achieved state-of-the-art results in image captioning. The deep architecture that these two papers suggest uses a pre-trained CNN such as AlexNet~\cite{NIPS2012_4824} or VGG~\cite{DBLP:journals/corr/SimonyanZ14a} to extract the features from the image that is captioned. The activation layer from the pre-trained network is then used as an input feature in the caption generator. The caption generator uses a language model to model the captions in the form of a vanilla recurrent neural network or a long-short term memory recurrent neural network. These architectures model caption generation as a process of predicting the next word in a sequence. 
Other significant research~\cite{DBLP:journals/corr/AntolALMBZP15},~\cite{DBLP:journals/corr/RenHG015},~\cite{DBLP:journals/corr/GaoMZHWX15},~\cite{DBLP:journals/corr/MalinowskiF14} has been done in image-based question answering. Datasets have been introduced by~\cite{DBLP:journals/corr/MalinowskiF14} and~\cite{DBLP:journals/corr/AntolALMBZP15} where images have been combined with question-answer pairs. The proposed solutions for this problem are similar to image captioning.  They have only an additional RNN for modeling the question. The combination of image features from the CNN and the question features extracted with RNN are used as an input in the RNN that generates the answers.
\subsection{Description of images-in-sequence}
Description of images-in-sequence covers the problems that are related to generating a description about a sequence of images or a video. The main focus of this type of multi-frames to sentence modelling is to capture the temporal dynamics of an image sequence and map them to a variable-length of words. There is no benchmark dataset for this type of caption generation, but some of the most frequently used datasets are the Youtube2Text dataset~\cite{guadarrama:iccv13}, Microsoft Research Video Description Corpus~\cite{collecting-highly-parallel-data-for-paraphrase-evaluation}, the movie description datasets M-VAD~\cite{DBLP:journals/corr/TorabiPLC15}, MPII-MD~\cite{DBLP:journals/corr/RohrbachRTS15} and the UCF101 Dataset~\cite{DBLP:journals/corr/abs-1212-0402}. A common solution is a sequence to sequence modelling, where a pre-trained CNN is used for feature extraction from the images (that are part of the sequence) and an RNN is used to model the temporal behaviour of the sequence of image features. The approach proposed by Yao et al.~\cite{1502.08029} employs a 3D CNN to extract local action features from every image in the sequence and an attention-based LSTM to exploit the global structure of the sequence. 

\subsection{Stories for images-in-sequence}
Stories for images-in-sequence explore the task of image streams to sentence sequences. 
Park et al.~\cite{NIPS2015_5776} tackle the problem of describing sequences of images with more narrative language. They introduce the NY (New York) and Disney datasets that they obtained from a vast user-generated resource of blog posts as text-image data. The blogs-posts were about people's experiences while visiting New York and Disneyland.  Every image from the blog-post is followed by a really long story that may or may not include information about the visual context of the image. This is the problem of these two datasets and that is why Huang et al.~\cite{DBLP:journals/corr/HuangFMMADGHKBZ16} introduce the VIST (Visual Storytelling Dataset). VIST is better then the aforementioned NY and Disney datasets, because every image in the image sequence is paired with one sentence from the story. 
The baseline approach is based on a Sequence to Sequence model that encodes the image features (extracted using a pre-trained CNN) with a GRU recurrent neural network~\cite{DBLP:journals/corr/ChungGCB14}. In their work, they encode 5 images and try to learn the 5 sentences that are associated with them all together. Other work that uses the VIST dataset is given by Yu et al.~\cite{DBLP:journals/corr/abs-1708-02977} and Liu et al.~\cite{let-photos-talk-generating-narrative-paragraph-photo-stream-via-bidirectional-attention-recurrent-neural-networks}. In~\cite{DBLP:journals/corr/abs-1708-02977}, the authors propose a model composed of three hierarchically-attentive Recurrent Neural Networks to encode the album photos, select representative (summary) photos and compose the story. On the other hand in~\cite{let-photos-talk-generating-narrative-paragraph-photo-stream-via-bidirectional-attention-recurrent-neural-networks} they propose a solution where the model learns a semantic space by jointly embedding each photo with its corresponding contextual sentence/ 
They present a novel Bidirectional Attention-based Recurrent Neural Network (BARNN) model, which can attend on the discovered semantic relation to produce a sentence sequence and maintain its consistency with the photostream.

\section{Dataset}

The dataset we used to train and test our model is the Visual Storytelling Dataset (VIST). VIST consists of 210,819 unique photos and 50,000 stories. The images were collected from albums on Flickr, using Flickr API.3. The albums included 10 to 50 images and all the images in an album are taken in a 48-hour span. This enables the dataset to have \say{storyable} images. The stories were created by workers on Amazon Mechanical Turk, where the workers were instructed to choose five images from the album and write a story about them. Every story has five sentence-stories and every sentence-story is paired with its appropriate image.
The dataset is split into 3 subsets, a training set (80\%), a validation set (10\%) and a test set (10\%). All the words and interpunction signs in the stories are separated by a space character and all the location names are replaced with the word location. Also, all the names of people are replaced with the words male or female depending on the gender of the person. One of the problems is that the stories were created by people, so not all stories necessarily have a story flow and from our overview of the dataset, some stories are not even correlated with the sequence of images. Because of this, we do not expect perfect stories in our results, but we want our model to generate stories with narrative language. We also expect the stories to contain words that will describe the visual context of the image in the sequence. 

\section{Architecture of proposed solution}

In order to model storytelling with narrative and visual components accurately, we should consider the human observation of creating stories for a sequence of images. When we see the first image, we start the story with a sentence that describes and evaluates the context of that particular image. For the next image in the sequence, we analyze the current image but we also consider the influence of the previous image because that's the only way we can preserve the temporal dependencies between the events in the images. Therefore for every image in the sequence, we consider the images that have happened in the past. Besides the previous images, it is important to preserve the temporal dependencies between the sentence-story generated for the previous image in the sequence (previous sentence-story) and the sentence-story generated for the current image in the sequence (current sentence-story). We achieve better story flow by considering the previous sentence-story, while we generate the sentence-story of the current image. Dissecting the way humans create stories, helped us conclude that the problem of generating stories for a sequence of images using visual and narrative components comes down to the way we model the sequence of images and the previous sentence-story.


The architecture that we propose in this paper is based on the Sequence to Sequence model~\cite{DBLP:journals/corr/SutskeverVL14} described in the previous sections. It incorporates encoder and decoder modules in the same fashion as the referred model. As a novelty, our encoder module is composed of two separate encoders, one that models the behaviour of the image sequence and other that models the sentence-story generated for the previous image in the sequence of images. 


\begin{figure}
\centering
\includegraphics[width=1.0\linewidth]{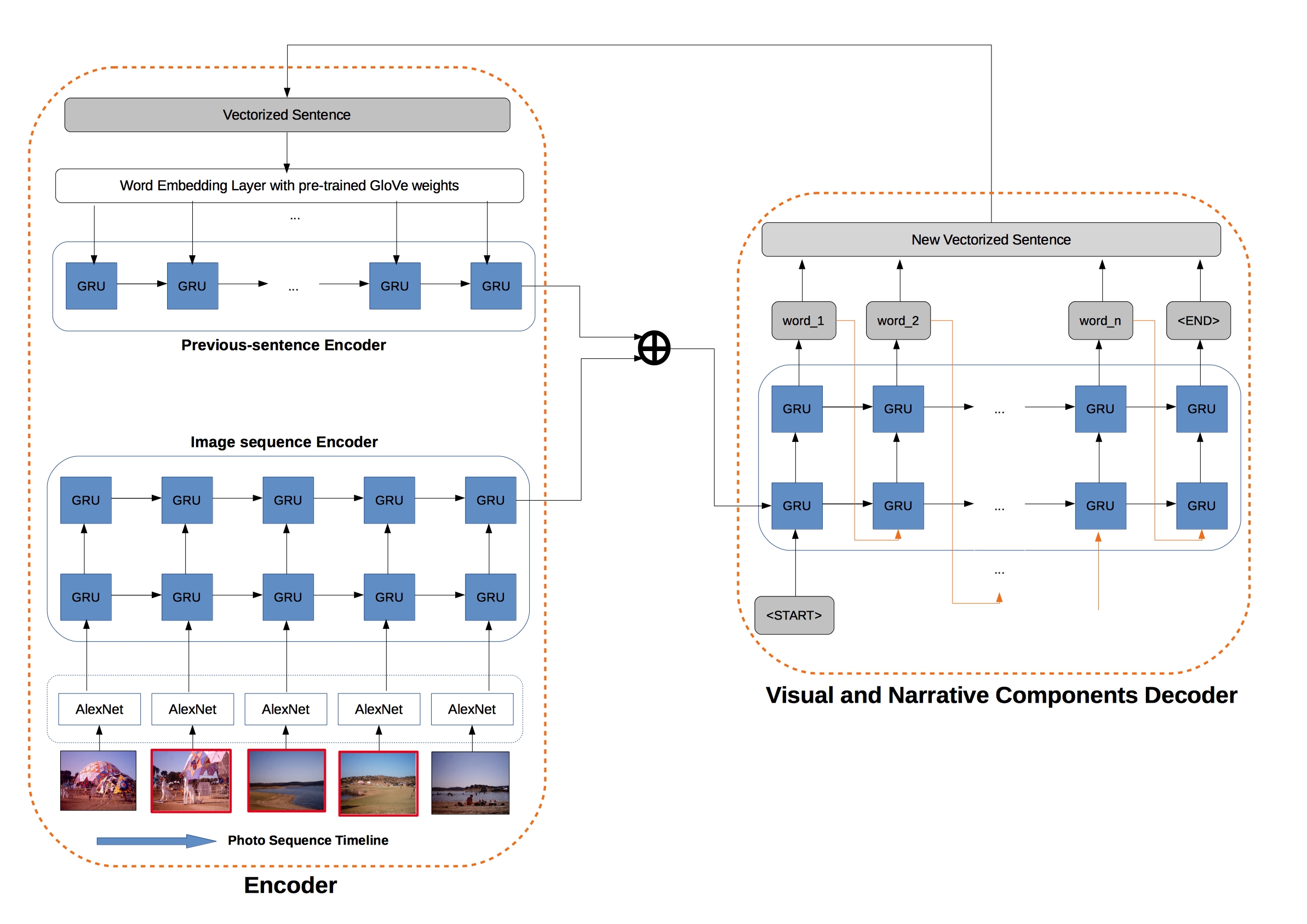}
\caption{The architecture of the proposed model. The images highlighted with red are the ones that are encoded and together with the previous sentence, they influence the generated sentence in the current time step.} \label{fig1}
\end{figure}

Recurrent neural networks have been very successful in sequence modelling because they can learn temporal dependencies between the elements of sequential data and it has been proven that they are appropriate for modelling a sequence of image features vectors. The encoder that we propose for modelling the behaviour of the image sequence aligns every sentence from the story with a sequence of images. This means that a sentence-story is generated per image while considering an appropriate number of images from the sequence. Opposite to our solution, the authors of the VIST dataset~\cite{DBLP:journals/corr/HuangFMMADGHKBZ16} propose aligning of the image sequence with the entire story. This means that after the story is generated, it should be divided into sentence-stories and each sentence-story should be assigned to a particular image from the sequence, which could be a drawback. 

The previous sentence-story encoder is also a recurrent neural network that learns the temporal behaviour of words in the sentence-story generated for the previous image. The two encoders of the proposed solution produce two fixed-length vector representation, one for the image-sequence and another for the previous sentence-story. In order to create a joint representation of the two encoders, we concatenate the vector representations. The concatenated vector is used as an initial hidden state of the decoder. In this way, we condition the decoder with the vector representations from the encoder module. The decoder module is a recurrent neural network that \say{translates} what the encoder module has produced. 

During the training process, we feed the encoder model with the image sequences and the previous sentence-stories, while the story decoder model with the current sentence-stories. When it's time to generate a story, we feed the encoder model with an image sequence, a previous sentence-story and the decoder model with the \textless START\textgreater ~token. After the decoder generates a word, that word is the input in the next time step of the decoder. When we generate the whole sentence-story, we append it to the story and use it as the previous sentence-story for the next generative process. The architecture of the proposed solution can be seen in Figure.~\ref{fig1}. The complete code and documentation of this project can be found on github\footnote{\url{https://github.com/Pendulibrium/ai-visual-storytelling-seq2seq}}.

\section{Experimental setup}

We used the $fc7$ vectors from the AlexNet convolutional neural network~\cite{NIPS2012_4824} to describe the images. We chose AlexNet over other more precise convolutional neural networks because AlexNet is less computationally expensive than other deeper networks. First, we transformed every image from RGB to BGR and after that, we re-sized the images with respect to their ratio. Also, we cropped them centrally to fit the dimensions of the input layer of AlexNet, because we assumed that the important information in the image is placed in its centre. 

The vocabulary that we created is composed of the most frequent words (words that appear at least 4 times in the stories). Also, we added \textless NULL\textgreater , \textless START\textgreater, \textless END\textgreater ~and \textless UNK\textgreater ~tokens to the vocabulary. After creating the vocabulary we decided that all the sentences would have a length of 20 words. We chose this number because most of the sentences had a length of 3 or 20 words. This meant that we would limit the longer sentences to 20 words and fill the shorter sentences with \textless NULL\textgreater token. We added the \textless START\textgreater token in front of every sentence and added the \textless END\textgreater token at their end. Every word that appears less than 4 times was substituted with the \textless UNK\textgreater token.
 
Before the sentence vectors entered the previous sentence-story encoder and the current sentence-story decoder, they passed through an embedding layer. The embedding layer used pre-trained word vectors obtained from the GloVe model~\cite{pennington2014glove}. This transformed the sentence, from a vector of 22 words (two words for the \textless START\textgreater and \textless END\textgreater tokens) to a vector of 22-word representations.
 
We experimented with an LSTM~\cite{Hochreiter:1997:LSM:1246443.1246450} and GRU~\cite{DBLP:journals/corr/ChoMBB14} recurrent network. There was no difference in the results, but because GRUs are less computationally expensive our solution uses GRUs. We also used two stacked GRUs together for the image sequence encoder and decoder, because stacking recurrent neural networks helps us model more complex sequences. Recurrent neural networks are inherently deep in time since their hidden state is a function of all previous hidden states, but they benefit from increasing their depth in space just like conventional deep networks do from stacking feedforward layers~\cite{DBLP:journals/corr/abs-1303-5778}. 

After various experiments with the size of the the GRU units for the components, we concluded that the best results were achieved when the GRU units for the image sequence encoder have 1024 neurons and the GRU unit for the previous-sentence encoder has 512 neurons. The encoders were set up in this way because it allowed the image sequence to have more impact on the generated sentence. Because of the concatenation of the outputs of the encoders, the GRU units for the decoder have 1536 neurons. 

Categorical cross entropy was used as a loss function, because it is the preffered loss function in Neural Machine Translation. The learning rate was set to 0.0001, because with greater learning rate our network was unable to imporove during training. Adam algorithm was used as an optimization algorithm. Adam algorithm is computationally more efficient than stochastic gradient descent, has little memory requirements, and it is invariant to the diagonal rescaling of the gradients. Also, it is well suited for problems that are large in terms of data and/or parameters~\cite{DBLP:journals/corr/KingmaB14}. 

To reduce the overfitting of the neural network during the training process, we used dropout as regularization. After experimenting with dropout on the input layers and the layers within the recurrent neural networks, we achieved best results when we applied dropout of 0.3 on the input layer and 0.5 on the layer before the softmax layer. The last parameter we had to choose was how many images in the past we will consider given the current image. When we trained the model with all the previous images, the last sentence-story always represented a summarization of all the images in the sequence and that resulted in a very generic sentence that didn't give any sufficient information. The best results were obtained when we considered the last three images.

\subsection{Evaluation metrics}

For evaluation of our generated stories, we used BLEU and METEOR score. These two metrics are usually used for evaluating Neural Machine Translation. The BLEU metric is designed to measure how close a generated translation is to that of human reference translations. In our case, it measures how close a generated story is to the original. It is important to note that stories, generated or original, may differ significantly in word usage, word order, and phrase length. To address these complexities, BLEU attempts to match variable length phrases between the generated story and the original story .
The METEOR method uses a sophisticated and incremental word alignment method that starts by considering exact word-to-word matches, word stem matches, and synonym matches. Alternative word order similarities are then evaluated based on those matches. 
These measures use a scale from 0 to 100 to quantify how similar the generated story is to the original based on a mechanical analysis of how many of the same words show up and how likely they are to appear in the same order.
It has also been shown that a high score (as a result of a method
which uses n-grams) probably indicates a good generation but a low score is
not necessarily an indication of a poor generation. This was one of the major problems that we faced when evaluating our models. In reality, two people can create very different stories about the same sequence of images, and both stories can be valid because perception is subjective.

\section{Results and analysis}

\subsubsection{Results and quantitative analysis}

In order to find the optimal solution, during the training of the models we did a quantitative analysis of the generated stories. The quantitative analysis was done by tracking the BLEU and METEOR scores of the trained models. In Table~\ref{tab1} we have presented the scores for three models (model1, model2, model3) that have the same network configuration (described in the previous section). The only difference between them is the number of epochs used for their training. After numerous experiments with the number of epochs used for training, we concluded that the model achieves the best scores when after training the loss (calculated over the training set) is between 0.82 and 1.72. Model1 has been trained for 50 epochs and has the smallest training loss, but it achieved the worst BLEU and METEOR scores out of the three models. This means that model1 is overfitting. Both model2 and model3 have a METEOR score of 23.9, but model3 achieved the best BLEU score. For comparison the baseline model provided in~\cite{DBLP:journals/corr/HuangFMMADGHKBZ16} achieved a METEOR score of 27.76. The difference in BLEU and METEOR scores between the model2 and model3 is very small and because of the aforementioned problems regarding the use of BLEU and METEOR scores as metrics for story generation, the results from the models had to be evaluated by human evaluators.
\begin{table}
\caption{Results for the generated stories. The loss is calculated over the training set and the METEOR and BLEU score are calucated over the test set.}\label{tab1}
\begin{tabular}{|l|l|l|l|}
\hline
Models &  $model 1$ & $model 2$ & $model 3$ \\
\hline
Training loss & 0.82 & 1.01 & 1.72\\
\hline
Number of epochs &  50 & 30 & 19\\
\hline
BLEU score & 24.5/9.0/3.2/1.3 & 26.0/9.7/3.6/1.5 & 26.4/10.1/3.8/1.6\\
\hline
METEOR score & 23.0 & 23.9 & 23.9\\
\hline
\end{tabular}
\end{table}

\begin{figure}
\centering
\includegraphics[width=0.9\linewidth]{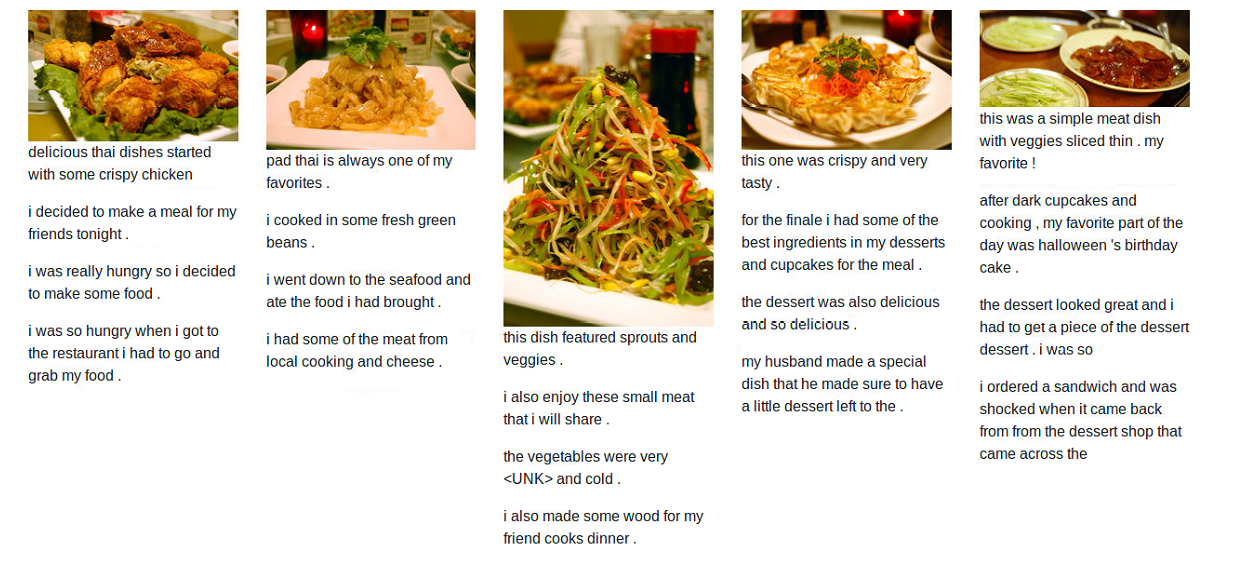}
\includegraphics[width=0.9\linewidth]{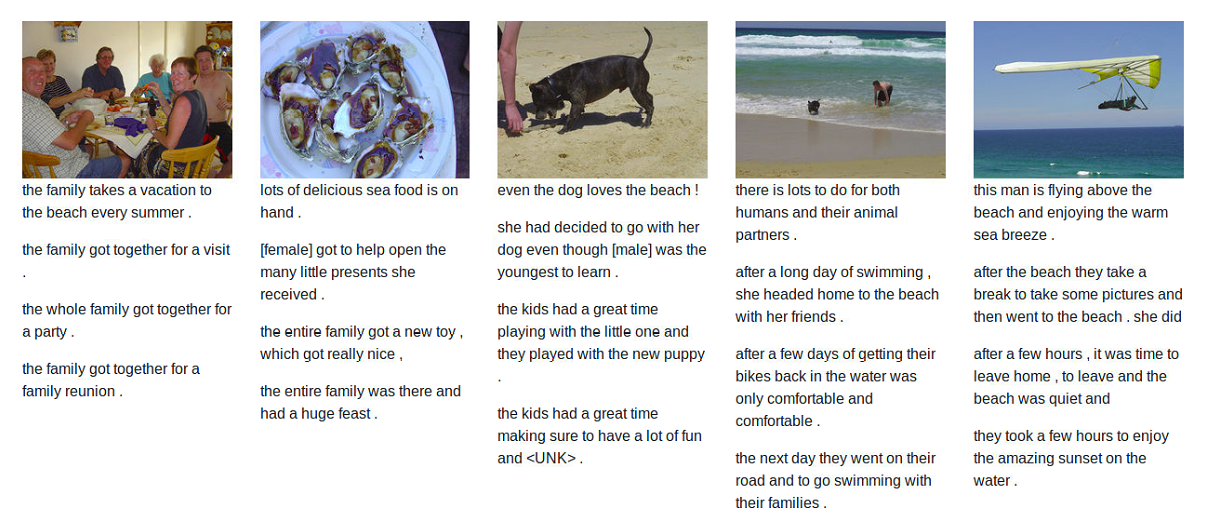}
\includegraphics[width=0.9\linewidth]{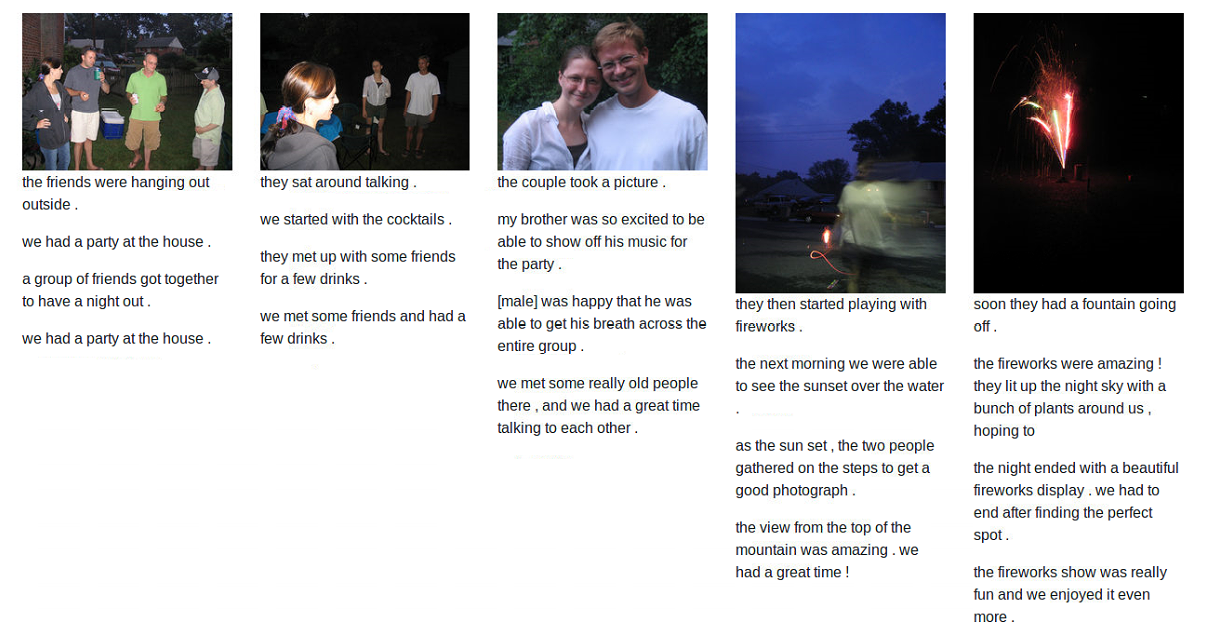}
\caption{In this figure, we can see the generated stories from the aforementioned models. The first row represents the original story, the second, third and fourth row are the generated story from the models respectively.} \label{fig2}
\end{figure}

\subsubsection{Qualitative analysis}
The human evaluation was done by the authors personally. From reading the generated stories and analyzing how they associate with their corresponding pictures we concluded that the best results were obtained by the model with loss of 1.01. We came to this conclusion because the generated stories by this model were better than the other models in terms of story flow and story length. Moreover, the generated stories from this model contained more words that described the visual context of the image sequence. The model with loss of 0.82 generated stories that had a lot of grammatical mistakes in them and we think that is happening because the model has over-fitted the training data. The model with a loss of 1.72 produced similar stories to the stories from the model with loss of 1.01, but it was slightly worse when it came about generating words that described the visual context of the image sequences.

Figure~\ref{fig2} shows the generated stories from the three models, for a given image sequence. More images with generated stories from the three models can be seen on github~\footnote{\url{https://github.com/Pendulibrium/ai-visual-storytelling-seq2seq/tree/master/results/images}}.

\section{Conclusion}
After a lot of experiments, we can conclude that the results from our proposed solution satisfied our expectations. The image-sequence encoder successfully learned the dependencies between the images and the proposed architecture was able to model the complex relations between the images and the stories. The improved story flow is a result of the inclusion of the previous sentence-story encoder. This encoder also contributed to the increase in the length of the generated stories. From the quantitative evaluation, it was obvious that metrics such as BLEU and METEOR are good for a distinction between really bad and supposedly good models. With the help of human evaluation, we concluded that most of the generated stories from our model made sense and looked like a story a human would tell. In order to improve our solution, in the future, we will focus on 3D convolutional neural networks for modelling the image sequences. Also, we will focus on the use of attention based models, because they will produce better alignment between the previous-sentence encoder and the decoder in our architecture. 




%
%
%
\bibliographystyle{splncs04}
\bibliography{mybibliography}
\end{document}